\documentclass[letterpaper]{IEEEtran}
\usepackage{latexsym,,amssymb,amsmath,graphicx,epsf,cite,bbm,float}
\usepackage{ifpdf}
\usepackage{epstopdf}

\usepackage{algorithm,algorithmic}
\usepackage{amsmath,amssymb,bm}
\usepackage{amsfonts,dsfont,color,bbm,subcaption}

\usepackage[linguistics]{forest}

\newcommand{\be}{\begin{equation}}
\newcommand{\ee}{\end{equation}}
\newcommand{\bea}{\begin{eqnarray}}
\newcommand{\eea}{\end{eqnarray}}

\newcommand{\MB}{\left[\begin{array}}
\newcommand{\ME}{\end{array}\right]}

\newcommand{\ei}{\end{itemize}}
\newcommand{\bi}{\begin{itemize}}

\newcommand{\abs}[1]{|#1|}

\usepackage{amsmath,amsthm}
\usepackage{hyperref}

\DeclareMathOperator*{\argmax}{arg\,max}
\DeclareMathOperator*{\argmin}{arg\,min}

\newtheorem{theorem}{Theorem}

\newtheorem{example}[]{Example}

\newtheorem{lemma}[]{Lemma}

\newtheorem{proposition}[]{Proposition}

\newtheorem{remark}[]{Remark}

\newtheorem{definition}[]{Definition}

\newtheorem{assumption}[]{Assumption}

\begin{document}

\title{Efficient, Anytime Algorithms for Calibration with Isotonic Regression under Strictly Convex Losses} 
\author{
	\IEEEauthorblockN{Kaan Gokcesu}, \IEEEauthorblockN{Hakan Gokcesu}
}
\maketitle

\flushbottom
\begin{abstract}
	We investigate the calibration of estimations to increase performance with an optimal monotone transform on the estimator outputs. We start by studying the traditional square error setting with its weighted variant and show that the optimal monotone transform is in the form of a unique staircase function. We further show that this staircase behavior is preserved for general strictly convex loss functions. Their optimal monotone transforms are also unique, i.e., there exist a single staircase transform that achieves the minimum loss. We propose a linear time and space algorithm that can find such optimal transforms for specific loss settings. Our algorithm has an online implementation where the optimal transform for the samples observed so far are found in linear space and amortized time when the samples arrive in an ordered fashion. We also extend our results to cases where the functions are not trivial to individually optimize and propose an anytime algorithm, which has linear space and pseudo-linearithmic time complexity.
\end{abstract}

\section{Introduction}
	\subsection{Calibration}
	In the problems of learning, recognition, estimation or prediction \cite{poor_book, cesa_book,russel2010}; decisions are often produced to minimize certain loss functions using features of the observations, which are generally noisy, random or even missing. There are numerous applications in a number of varying fields such as decision theory \cite{tnnls4}, control theory \cite{tnnls3}, game theory \cite{tnnls1, chang}, optimization \cite{zinkevich, hazan}, density estimation and anomaly detection \cite{gokcesu2018density, willems, gokcesu2018anomaly,coding2,gokcesu2019outlier, gokcesu2016nested}, scheduling \cite{kose2020novel}, signal processing \cite{ozkan,gokcesu2018semg}, forecasting \cite{singer, gokcesu2016prediction} and bandits \cite{neyshabouri2018asymptotically,cesa-bianchi,gokcesu2018bandit}. These decisions are acquired from specific learning models, where the goal is to distinguish certain data patterns and provide accurate estimations for practical use. Most learning methods produce estimators that return some sort of score, which can be used to rank them in a given problem scenario (e.g. membership probabilities). However, in many applications, the ranking (ordinality) by itself is not sufficient and precise estimates may be required. To this end, it has become important to develop post-processing calibration methods \cite{lichtenstein1977calibration,keren1991calibration}. When calibration is done accordingly, the estimates should match better with the targets (i.e., minimize some sort of loss function); however, deviations from perfect calibration are common in practice and vary depending on the estimation models \cite{naeini2015obtaining}. Well-calibrated estimations are required in many areas of science, medicine and business. It is of crucial importance not only for the decision making process, but also to compare different estimators \cite{zhang2004naive, jiang2005learning, hashemi2010application}, or to combine them \cite{bella2013effect}. Although research on calibration is not as extensive as high discrimination learning models; there are several methods to create well-calibrated estimations, which require a form of regularization to avoid over-fitting \cite{gokcesu2021optimally}. 
	
	\subsection{Parametric Mapping}
	We can straightforwardly regularize the calibration mapping by fitting a parametric shape (or function). An example is the fitting of a step function on the binary classifiers where the parameter is the threshold \cite{gokcesu2021optimally}. Another popular approach is the sigmoidal fitting \cite{platt1999probabilistic}, where the scores are mapping to the calibrated estimations by using a sigmoid function. There various methods to tune the sigmoid parameters by minimizing the cumulative loss \cite{gill2019practical}.
	Traditionally, these approaches are developed to transform the score outputs of certain learning models into more meaningful estimations such as class predictions or membership probabilities (empirically, the sigmoid mapping is shown to produce accurate estimations). In spite of the higher efficiency of the sigmoidal fitting compared to simply training another post processing learning model, it has been empirically shown that the performance does not suffer for many datasets. Although the work in \cite{platt1999probabilistic} studied the SVM learners, it has also been utilized to calibrate other learners as well \cite{niculescu2005predicting,bennett2000assessing}. However, it has also been empirically shown that the sigmoid transform does not always have accurate calibrations for different learning models \cite{zadrozny2002transforming}. The reason is the inherent restrictiveness of the parametric shape \cite{jiang2012calibrating}, which may be undesirable despite its computational efficiency.
	
	\subsection{Quantile Binning}
	In the quantile binning approach \cite{zadrozny2001learning, zadrozny2001obtaining}, we start by first ordering the score outputs of a learner and partition them in equal sized frequency bins. Then, the calibrated estimations are created from the score outputs in the respective bins individually, which results in an inherent regularization on the calibrated estimations, where we enforce the mapping of similar scores to similar values. Although, in certain problem settings, the quantile binning may coincide with the parametric mapping \cite{gokcesu2021optimally}; it is generally a less restrictive approach. However, despite its efficiency and milder restrictions, the bin sizes and locations remain fixed over time and are hard to optimize \cite{zadrozny2002transforming}. If the number of bins, their sizes and locations are not determined accurately, it may produce undesirable (or inaccurate) calibrated estimations. To address this issue, the work in \cite{jiang2012calibrating} produces a confidence interval for each estimation to build its frequency bins. In \cite{naeini2015obtaining}, the authors address its limitations by creating different equal frequency binning models and their aggregation \cite{heckerman1995learning}. The work in \cite{naeini2015binary} address this drawback by Bayesian combination of all possible binning models inferred from the data samples, which is unfortunately inefficient. Nevertheless, the relaxed restriction may become detrimental since certain information from the score outputs are not fully utilized.
	
	\subsection{Isotonic Regression}
	In the calibration of score outputs, the shortcomings of parametric mapping and quantile binning have been addressed by the non-parametric approach of isotonic regression \cite{robertson1988order,zadrozny2002transforming}. It has an inherent regularity, which comes from the monotonic mapping from the uncalibrated score outputs of a learner to the calibrated estimations, i.e., its mapping enforces the score outputs' own ordinality to the calibrated estimations. Because of this restriction, we end up with several adjacent samples mapping to a single value, which is where the name isotonic comes from. This mapping can directly find the optimal calibrations for each sample as long as the original and calibrated scores have matching ordinality. In the event of mismatch, the calibrator aggregates the adjacent samples. The number of samples used to determine a single mapping increase as the ordinality mismatch becomes worse. Henceforth, it is similar to a quantile binning strategy, where the number of bins, sizes and locations are derived from the initial monotone regularity \cite{zadrozny2002transforming}. Moreover, since it produces a monotone mapping, it is somewhat similar to the parametric sigmoid fitting as well \cite{gokcesu2021optimally}. Therefore, it is an approach that is in between the parametric mapping and the quantile binning methods. There are various existing techniques to find such mappings \cite{ayer1955empirical,brunk1972statistical}. It has also been incorporated to assist the quantile binning as well \cite{naeini2016binary,tibshirani2011nearly}. There are strategies that use isotonic regression to aggregate multiple score outputs \cite{zhong2013accurate} and for specific losses like the ranking loss \cite{menon2012predicting}. It has also been utilized to create online calibrators \cite{gokcesu2021optimally}.

\subsection{Contributions and Organization}

The calibration should conform to some form of regularity; which is achieved by the function model in the parametric mapping, individual treatment of the sample groups in quantile binning and monotonicity in isotonic regression \cite{gokcesu2021optimally}. Although all of them somewhat avoid over-fitting, there have their shortcomings. While the parametric mappings are easy to optimize, they suffer from high restrictions. On the contrary, the quantile binning is less restrictive but harder to optimize since the number of bins, locations and sizes are not easy to select. Although isotonic regression addresses these shortcomings, the existing approaches have limited applicability and may computationally suffer in the event of hard to optimize losses. To this end, we propose efficient and anytime algorithms for general strictly convex loss functions. In \autoref{sec:square}, we show that the traditional square error and its weighted variant has an optimal unique calibrator (transform or mapping) in the form of a staircase function. In \autoref{sec:strict}, we show that this optimal unique staircase mapping is preserved for a large class of loss functions, specifically the strictly convex losses. In \autoref{sec:offline}, we show how this unique optimal staircase can be found in linear time for certain types of losses. In \autoref{sec:online}, we propose an online approach that can find the optimal mapping when the output estimations arrive in order. In \autoref{sec:anytime}, we extend our algorithm to find mappings for general strictly convex loss functions, where the minimizer is not trivial to determine unlike the square error. In \autoref{sec:conc}, we finish with some concluding remarks.

\section{Optimal Monotone Transform For Square Error is a Unique Staircase Function}\label{sec:square}
In this section, we show that, for the regression problem of minimizing the square error, an optimal monotone transform on the score values produced by an estimator is a staircase function, where each stair value is given by the mean of the samples on that stair. We start with the formal problem definition similarly with \cite{gokcesu2021optimally}.
\subsection{Problem Definition}
	We have $N$ number of samples indexed by $n\in\{1,\ldots,N\}$. For every $n$, 
	\begin{enumerate}
		\item We have the value $y_n$, which is the target variable of the $n^{th}$ sample, i.e., 
		\begin{align}
		y_n\in\Re.\label{eq:y}
		\end{align}
		\item We have the output $x_n$ of an estimator, which is the score of the $n^{th}$ sample, i.e.,
		\begin{align}
		x_n\in\overline\Re.\label{eq:x}
		\end{align} 
		\item We map the score outputs $x_n$ of the estimator to nondecreasing variables $z_n$ with a mapping function $C(\cdot)$, i.e., 
		\begin{align}
		z_n\triangleq C(x_n)\in\Re.\label{eq:p}
		\end{align}
		\item We choose the mapping $C(\cdot)$ such that it is a monotone transform (monotonically nondecreasing), i.e.,
		\begin{align}
		C(x_n)\geq C(x_{n'}) \text{ if } x_n> x_{n'}.\label{eq:C}
		\end{align} 
	\end{enumerate}
	For the given setting above, we have the following problem definition for the minimization of the square error regression.
	\begin{definition}\label{def:problem}
		For a given set of score outputs $\{x_n\}_{n=1}^N$ and target values $\{y_n\}_{n=1}^N$, the minimization of the square error is given by
		\begin{align*}
		\argmin_{C(\cdot)\in\Omega}\sum_{n=1}^{N}(C(x_n)-y_n)^2,
		\end{align*}
		where $\Omega$ is the class of all univariate monotonically nondecreasing functions that map from $\overline\Re$ to $\overline\Re$.
	\end{definition}
	The problem in \autoref{def:problem} aims to minimize the square regression error unlike the linear loss in \cite{gokcesu2021optimally}. Without loss of generality and problem definition, we make the following assumptions.
	\begin{assumption}\label{ass:monotone}
	Let $\{x_n\}_{n=1}^N$ be in an ascending order, i.e.,
	\begin{align*}
	x_{n-1}\leq x_n,
	\end{align*}
	since if they are not, we can simply order the scores $x_n$ and acquire the corresponding $y_n$ target variables.
	\end{assumption}
	
	Next, we show why the optimal monotone transform $C(\cdot)$ on the scores $x_n$ is a staircase function.
	
	\subsection{Optimality of Staircase for Square Error}\label{sec:optimal}
	Let us have a minimizer $C^*(\cdot)$ for \autoref{def:problem} given by
	\begin{align}
		C^*(x_n)=z^*_n, &&n\in\{1,\ldots,N\},\label{eq:prob}
	\end{align} 
	where $z^*_n$ be the corresponding monotone mappings.
	\begin{lemma}\label{thm:sample}
		If $C^*(\cdot)$ is a minimizer for \autoref{def:problem}, then
		\begin{align*}
			z^*_{n}=z^*_{n+1}\text{ if }y_n\geq y_{n+1}
		\end{align*}
	\begin{proof}
		By design, any monotone transform has $z^*_n\leq z^*_{n+1}$. Let the loss of interest be $L_n(z^*_n,z^*_{n+1})=(z^*_n-y_n)^2+(z^*_{n+1}-y_{n+1})^2$. Given $y_n\geq y_{n+1}$, we have the following six cases:
		\begin{enumerate}
			\item $z^*_n\leq z^*_{n+1}\leq y_{n+1}\leq y_n$:
			\subitem $L_n(z^*_{n+1},z^*_{n+1})\leq L_n(z^*_{n},z^*_{n+1})$
			\item $z^*_n\leq y_{n+1}\leq z^*_{n+1}\leq y_n$:
			\subitem $L_n(z^*_{n+1},z^*_{n+1})\leq L_n(z^*_{n},z^*_{n+1})$.
			\item $ z^*_n\leq y_{n+1}\leq y_n\leq z^*_{n+1}$:
			\subitem $L_n(\theta_n,\theta_n)\leq L_n(z^*_{n},z^*_{n+1})$, $\forall\theta_n\in[y_{n+1},y_n]$.
			\item $y_{n+1}\leq z^*_n\leq z^*_{n+1}\leq  y_n$:
			\subitem $L_n(\theta_n,\theta_n)\leq L_n(z^*_{n},z^*_{n+1})$, $\forall\theta_n\in[z^*_{n},z^*_{n+1}]$.
			\item $y_{n+1}\leq z^*_n\leq y_n\leq z^*_{n+1}$:
			\subitem $L_n(z^*_{n},z^*_{n})\leq L_n(z^*_{n},z^*_{n+1})$.
			\item $y_{n+1}\leq y_n\leq z^*_n\leq z^*_{n+1}$:
			\subitem $L_n(z^*_{n},z^*_{n})\leq L_n(z^*_{n},z^*_{n+1})$.
		\end{enumerate}
		Therefore, if $C^*(\cdot)$ is a minimizer, then $z^*_{n}=z^*_{n+1}$, which concludes the proof.
	\end{proof}
	\end{lemma}

	Hence, there are groups of samples with the same mapping $z^*_n$. Following \cite{gokcesu2021optimally}, let there be $I$ groups, where the group $i\in\{1,\ldots,I\}$ cover the samples $n\in\mathcal{N}_i=  \{n_i+1,\ldots,n_{i+1}\}$ (where $n_1=0$ and $n_{I+1}=N$), i.e., for each group $i$, we have $z^*_n=z^*_{n'}, \forall n,n'\in\mathcal{N}_i$. With abuse of notation, let us denote the mapping of the $i^{th}$ group with $z^*_i$. We have the following result.
	
	\begin{lemma}\label{thm:group}
	If $C^*(\cdot)$ is a minimizer for \autoref{def:problem}, then
	\begin{align}
		z^*_{i}=z^*_{i+1}\text{ if }\tilde{y}_i\geq \tilde{y}_{i+1},
	\end{align}
	where $\tilde{y}_i$ is the mean of the target variables $y_n$ of group $i$, i.e., $\tilde{y}_i\triangleq \frac{1}{n_{i+1}-n_i}\sum_{n=n_i+1}^{n_{i+1}}y_{n}$. Consequently, we have
	\begin{align}
		z^*_{i}<z^*_{i+1} \text{ only if } \tilde{y}_i<\tilde{y}_{i+1},
	\end{align}
	i.e., the $i^{th}$ groups mapping $z_i^*$ is distinct (consequently strictly less) than $z^*_{i+1}$ only if $\tilde{y}_i$ is strictly less than $\tilde{y}_{i+1}$.
	\begin{proof}
		The proof follows from the proof of \autoref{thm:sample}.
	\end{proof}
	\end{lemma}
	
	\begin{theorem}\label{thm:staircase}
		If $C^*(\cdot)$ is a minimizer for \autoref{def:problem} with $I^*$ uniquely mapped groups (with distinct mappings $z^*_i$) such that
		\begin{align*}
			C^*(\{x_n\}_{n=n_i+1}^{n_{i+1}})=\tilde{y}_i&&|&&\tilde{y}_i<\tilde{y}_{i+1}, \forall i\in\{1,\ldots,I^*-1\},
		\end{align*}
		i.e., we have $z^*_i=\tilde{y_i}$.
		\begin{proof}
			The proof comes from \autoref{thm:group}, where we can directly set $z^*_i=\tilde{y}_i$ to minimize the square loss.
		\end{proof}
	\end{theorem}

	In this section, we have shown that there exists an optimal mapping $C^*(\cdot)$ (with monotone behavior) that minimizes the square error in \autoref{def:problem}, which is also a staircase function on the score variables $x_n$. In the next section, we show that a staircase function again minimizes a weighted version of the problem in \autoref{def:problem}.
	
	\subsection{Extension to Weighted Square Error}
	In this section, we prove that the optimal monotone transform that minimizes a weighted square error instead of the vanilla square error is again a staircase function. 
	The setting in \eqref{eq:y}, \eqref{eq:x}, \eqref{eq:p}, \eqref{eq:C} remains the same (much like \cite{gokcesu2021optimally}). We also make the same assumption in \autoref{ass:monotone}.
	The revised version of the problem in \autoref{def:problem} is as the following.
	\begin{definition}\label{def:problemW}
		For $\{x_n\}_{n=1}^N$, $\{y_n\}_{n=1}^N$, the minimization of the strictly convex error is given by
		\begin{align*}
		\argmin_{C(\cdot)\in\Omega}\sum_{n=1}^{N}\alpha_n (C(x_n)-y_n)^2,
		\end{align*}
		where $\Omega$ is the class of all univariate monotonically increasing functions that map from $\overline\Re$ to $\overline\Re$ and $\alpha_n>0$ is the sample weight of the error for $x_n$.
	\end{definition}
	Let us again assume that there exists a monotone transform $C^*(\cdot)$ that minimizes \autoref{def:problemW} with its corresponding mappings for each sample as in \eqref{eq:prob}.
	\autoref{thm:sample} directly holds true similarly and we end up with $I$ groups of consecutive samples (with the corresponding intervals) that map to the same value $z^*_i$. \autoref{thm:group} is modified as the following.
	
	\begin{lemma}\label{thm:groupW}
		If $C^*(\cdot)$ is a minimizer for \autoref{def:problemW}, then
		\begin{align}
			z^*_{i}=z^*_{i+1}\text{ if }\tilde{y}_i\geq \tilde{y}_{i+1},
		\end{align}
		where $\tilde{y}_{i}$ is the weighted mean, i.e.,
		\begin{align}
			\tilde{y}_{i}=\frac{\sum_{n=n_i+1}^{n_{i+1}}\alpha_n y_n}{\sum_{n=n_i+1}^{n_{i+1}}\alpha_n }.
		\end{align}
		Consequently, we have
		\begin{align}
			z^*_{i}<z^*_{i+1} \text{ only if } \tilde{y}_i<\tilde{y}_{i+1},
		\end{align}
		i.e., the $i^{th}$ groups mapping $z_i^*$ is distinct (or strictly less) than $z^*_{i+1}$ only if $\tilde{y}_i$ is strictly less than $\tilde{y}_{i+1}$, where $\tilde{y}_i$ is the weighted mean of the target variables of the group $i$.
		\begin{proof}
			The proof follows \autoref{thm:group} and the proof of \autoref{thm:sample} with the altered problem definition.
		\end{proof}
	\end{lemma}
	
	Hence, the comparison is done over the weighted mean of the groups instead of the simple mean. When $\alpha_n=1$ for all $n$, the problem definition and the result becomes equivalent to the preceding sections.
	\begin{theorem}\label{thm:staircaseW}
	If $C^*(\cdot)$ is a minimizer for \autoref{def:problemW} with $I^*$ uniquely mapped groups (with distinct mappings $z^*_i$) such that
	\begin{align*}
		C^*(\{x_n\}_{n=n_i+1}^{n_{i+1}})=\tilde{y}_i&&|&&\tilde{y}_i<\tilde{y}_{i+1}, \forall i\in\{1,\ldots,I^*-1\},
	\end{align*}
	i.e., we have $z^*_i=\tilde{y_i}$, where $\tilde{y}_i$ is the weighted mean.
	\begin{proof}
		The proof comes from \autoref{thm:groupW}. Since $z^*_{i}<z^*_{i+1}$ only if $\tilde{y}_i<\tilde{y}_{i+1}$, we can directly set $z^*_i=\tilde{y}_i$ to minimize the loss.
	\end{proof}
	\end{theorem}

		In this variant, we have shown that there exists an optimal mapping $C^*(\cdot)$ (with monotone behavior) that minimizes the weighted square error in \autoref{def:problemW}, which is also a staircase function on the score variables $x_n$. In the next section, we show that a staircase function again minimizes a generalized version of the problem in \autoref{def:problem} and \autoref{def:problemW}.
	
	\section{Optimal Monotone Transform for Strictly Convex Losses is a Staircase Function}\label{sec:strict}
	In this section, we generalize the results up to now to show that the optimal monotone transform on our estimations $x_n$ that minimizes general strictly convex losses is again a staircase function (similar with the extension to linear losses from classification error in \cite{gokcesu2021optimally}). 
	\subsection{Problem Definition}
	We again have samples indexed by  $n\in\{1,\ldots,N\}$, and for every $n$, 
	\begin{enumerate}
		\item We have the score output of an estimator
		\begin{align}
		x_n\in\overline\Re.\label{eq:xL}
		\end{align} 
		\item We have the strictly convex loss 
		\begin{align}
		l_n(\cdot)\label{eq:z}
		\end{align}
		\item We map these scores to values
		\begin{align}
		z_n\triangleq C(x_n)\in\Re.\label{eq:q}
		\end{align}
		\item The mapping $C(\cdot)$ is monotonically nondecreasing, i.e.,
		\begin{align}
		C(x_n)\geq C(x_{n'}) \text{ if } x_n> x_{n'}.\label{eq:CL}
		\end{align} 
	\end{enumerate}
	For this new setting, we have the following problem definition.
	\begin{definition}\label{def:problemL}
		For $\{x_n\}_{n=1}^N$, $\{l_n(\cdot)\}_{n=1}^N$, the minimization of the cumulative strictly convex loss is given by
		\begin{align*}
		\argmin_{C(\cdot)\in\Omega}\sum_{n=1}^{N}l_n(C(x_n)),\label{eq:problemL}
		\end{align*}
		where $\Omega$ is the class of all univariate monotonically increasing functions that map from $\overline\Re$ to $\overline\Re$.
	\end{definition}

	When $l_n(\cdot)$ is strictly convex, its second derivative on its argument is always positive, i.e.,
	\begin{align}
		\dfrac{\delta^2l_n(z)}{\delta z^2}\bigg\rvert_{z=z_0}>0, &&\forall z_0\in\Re
	\end{align} 
	Henceforth, the problem in \autoref{def:problemL} fully generalizes the problems in \autoref{def:problem} and \autoref{def:problemW}, since
	\begin{itemize}
		\item For \autoref{def:problem}, we have
		\begin{align*}
			l_n(z_n)=&(z_n-y_n)^2,
		\end{align*}
		and its second derivative is
		\begin{align*}
			\dfrac{\delta^2l_n(z)}{\delta z^2}=&2,
		\end{align*}
		which is positive, hence, strictly convex.
		\item For \autoref{def:problemW}, we have
		\begin{align*}
			l_n(z_n)=&\alpha_n(z_n-y_n)^2,
		\end{align*}
		and its second derivative is
		\begin{align*}
			\dfrac{\delta^2l_n(z)}{\delta z^2}=&2\alpha_n,
		\end{align*}
		which is positive, hence, strictly convex since $\alpha>0$ from \autoref{def:problemW}.
	\end{itemize}
	
	We make the same assumption as in \autoref{ass:monotone}. Next, we show why the optimal monotone transform $C(\cdot)$ on the scores $x_n$ is a staircase function for the general strictly convex losses.
	
	\subsection{Optimality of Staircase for Strictly Convex Losses}
	Let us again assume that there exists an optimal monotone transform $C^*(\cdot)$ in $\Omega$ that minimizes \autoref{def:problemL}, where $\Omega$ is the class of all univariate monotonically nondecreasing functions that map to the extended reals $\overline\Re$. Let the minimizer of $l_n(\cdot)$ be $y_n$, i.e., $y_n=\argmin_{y\in\overline{\Re}} l_n(y)$. We have a similar result to \autoref{thm:sample} as the following
	
	\begin{lemma}\label{thm:sampleL}
		If $C^*(\cdot)$ is a minimizer for \autoref{def:problemL}, then
		\begin{align}
			z^*_{n}=z^*_{n+1}\text{ if }y_n\geq y_{n+1},
		\end{align}
		where $y_n=\argmin l_n(y)$.
		\begin{proof}
			By design, any monotone transform has $z^*_n\leq z^*_{n+1}$. Let $L_n(z^*_n,z^*_{n+1})=l_n(z_n^*)+l_{n+1}(z^*_{n+1})$ be the loss of interest. Given $y_n\geq y_{n+1}$, we have the following six cases from the strict convexity:
			\begin{enumerate}
				\item $z^*_n\leq z^*_{n+1}\leq y_{n+1}\leq y_n$:
				\subitem $L_n(z^*_{n+1},z^*_{n+1})\leq L_n(z^*_{n},z^*_{n+1})$
				\item $z^*_n\leq y_{n+1}\leq z^*_{n+1}\leq y_n$:
				\subitem $L_n(z^*_{n+1},z^*_{n+1})\leq L_n(z^*_{n},z^*_{n+1})$.
				\item $ z^*_n\leq y_{n+1}\leq y_n\leq z^*_{n+1}$:
				\subitem $L_n(\theta_n,\theta_n)\leq L_n(z^*_{n},z^*_{n+1})$, $\forall\theta_n\in[y_{n+1},y_n]$.
				\item $y_{n+1}\leq z^*_n\leq z^*_{n+1}\leq  y_n$:
				\subitem $L_n(\theta_n,\theta_n)\leq L_n(z^*_{n},z^*_{n+1})$, $\forall\theta_n\in[z^*_{n},z^*_{n+1}]$.
				\item $y_{n+1}\leq z^*_n\leq y_n\leq z^*_{n+1}$:
				\subitem $L_n(z^*_{n},z^*_{n})\leq L_n(z^*_{n},z^*_{n+1})$.
				\item $y_{n+1}\leq y_n\leq z^*_n\leq z^*_{n+1}$:
				\subitem $L_n(z^*_{n},z^*_{n})\leq L_n(z^*_{n},z^*_{n+1})$.
			\end{enumerate}
			Therefore, if $C^*(\cdot)$ is a minimizer, then $z^*_{n}=z^*_{n+1}$, which concludes the proof.
		\end{proof}
	\end{lemma}

	We also have a result similar to \autoref{thm:group} as follows.
	\begin{lemma}\label{thm:groupL}
		If $C^*(\cdot)$ is a minimizer for \autoref{def:problemL}, then
		\begin{align}
			z^*_{i}=z^*_{i+1}\text{ if }\tilde{y}_i\geq \tilde{y}_{i+1},
		\end{align}
		where $\tilde{y}_i$ is the minimizer of the sum of the losses in group $i$, i.e., $\tilde{y}_i=\argmin\sum_{n=n_i+1}^{n_{i+1}}l_n(y)$. Consequently, we have
		\begin{align}
			z^*_{i}<z^*_{i+1} \text{ only if } \tilde{y}_i<\tilde{y}_{i+1},
		\end{align}
		i.e., the $i^{th}$ groups mapping $z_i^*$ is distinct (consequently strictly less) than $z^*_{i+1}$ only if $\tilde{y}_i$ is strictly less than $\tilde{y}_{i+1}$.
		\begin{proof}
			We already established that every sample in group $i$ maps to the same variable $z^*_i$. Thus, we can treat this group's total loss as a single sample. Since the sum of strictly convex losses is also a strictly convex loss, the proof follows from the proof of \autoref{thm:sampleL}.
		\end{proof}
	\end{lemma}

	Similar to \autoref{thm:staircase} and \autoref{thm:staircaseW}, we have the following theorem.
	
	\begin{theorem}\label{thm:staircaseL}
		If $C^*(\cdot)$ is a minimizer for \autoref{def:problemL} with $I^*$ uniquely mapped groups (with distinct mappings $z^*_i$) such that
		\begin{align*}
			C^*(\{x_n\}_{n=n_i+1}^{n_{i+1}})=\tilde{y}_i&&|&&\tilde{y}_i<\tilde{y}_{i+1}, \forall i\in\{1,\ldots,I^*-1\},
		\end{align*}
		i.e., we have $z^*_i=\tilde{y_i}$.
		\begin{proof}
			The proof comes from \autoref{thm:groupL}, where we can directly set $z^*_i=\tilde{y}_i$ to minimize the composite strictly convex loss.
		\end{proof}
	\end{theorem}

\section{Offline Algorithm to Find the Optimal Staircase Mapping for Strictly Convex Losses}\label{sec:offline}
In this section, we propose algorithms that can find an optimal staircase transform for the problem in \autoref{def:problemL}.

\subsection{Direct Approach}
The straightforward approach works as follows:

\begin{enumerate}
	\item At the beginning, let us have $N$ samples with the corresponding scores $\{x_n\}_{n=1}^N$ and losses $\{l_n(\cdot)\}_{n=1}^N$.
	\item Create a group for each sample, i.e., $n\in\{1,\ldots,N\}$. Set their group loss $L_n(\cdot)=l_n(\cdot)$. Set the initial minimizers $y_n=\argmin_y L_n(y)$ for all $n\in\{1,\ldots,N\}$ groups.
	\item Let $\mathcal{I}$ be the set of all indices $i$ such that the consecutive group pair $(i,i+1)$ have $y_i\geq y_{i+1}$.\label{step:check}
	\item IF $\mathcal{I}$ is not empty, join the groups $i$ and $i+1$ for all $i\in\mathcal{I}$, and re-index the new groups accordingly. If there is consecutive grouping, we join all of them together. For example, if we join the pairs $\{n,n+1\}$ and $\{n+1,n+2\}$, we join $\{n,n+1,n+2\}$. 
	We assign the new losses, where the joined groups' losses are summed together. We also calculate their new group minimizers $y_i$. Return to Step \ref{step:check}.
	\item ELSE (if $\mathcal{I}$ is empty), stop.
\end{enumerate}

Our algorithmic design stems from the notion of joining the groups in \cite{gokcesu2021optimally} together with the properties implied by \autoref{thm:sampleL} and \autoref{thm:groupL}. The first iteration is the result of the property of \autoref{thm:sampleL}, while the next iterations are the result of the more general property of \autoref{thm:groupL}.
Since we only keep track of the group losses $L_i(\cdot)$ and their minimizers $y_i$, we have a $O(T)$ space complexity. Let the optimal transform have $S$ steps in the staircase. This approach has at most $T-S$ combinations, which is $O(T)$. Thus, this approach can find an optimal transform in $O(T)$ space and $O(T\tau)$ time, where $\tau$ is the worst-case time it takes to find a minimizer. 

\begin{remark}
	For the weighted square errors in \autoref{sec:square}, it takes at most $O(T)$ time (i.e., $\tau=T$) to find the weighted group means. Thus, this approach can find an optimal transform in $O(T)$ space and $O(T^2)$ time.
\end{remark}

\begin{example}\label{example}
	Let $x_n=n$, $\alpha_n=1$ and $y_n$ be from
	\begin{align*}
	\{{44},{52},{18},{14},{93},{37},{96},{08},{01},{95},{21},{77},{46},{36},{69}\}
	\end{align*}
	for $n\in\{1,\ldots,N\}$, where $N=15$. Let the losses be $l_n(C(x_n))=(C(x_n)-y_n)^2$, i.e., the square loss.
	The iterations work as follows:
	\begin{enumerate}
		\item We start our algorithm with the single sample groups:  $\underline{44},\underline{52},\underline{18},\underline{14},\underline{93},\underline{37},\underline{96},\underline{08},\underline{01},\underline{95},\underline{21},\underline{77},\underline{46},\underline{36},\underline{69}$
		
		\item Then, we combine the groups according to the algorithm $\underline{44},\underline{52,18,14},\underline{93,37},\underline{96,08,01},\underline{95,21},\underline{77,46,36},\underline{69}$
		
		\item We calculate the minimizers of the newly joined groups $\underline{44},\underline{28,28,28},\underline{65,65},\underline{35,35,35},\underline{58,58},\underline{53,53,53},\underline{69}$
		\item We again combine the groups according to the algorithm $\underline{44,52,18,14},\underline{93,37,96,08,01},\underline{95,21,77,46,36},\underline{69}$
		\item We again calculate the minimizers of the joined groups $\underline{32,32,32,32},\underline{47,47,47,47,47},\underline{55,55,55,55,55},\underline{69}$
		\item Since there are no more groups to combine, the algorithm stops and we reach the optimal staircase mapping.
	\end{enumerate}
\end{example}

\subsection{Efficient Implementation}
We observe that although the memory usage is efficient, there is room for improvement in the time complexity based on the problem setting. Since much of the time complexity stems from the newly calculated group minimizers, an efficient update can speed up the algorithm. 

Let $\{y_i,y_{i+1},\ldots,y_{i+j}\}$ be the minimizers for the group losses $\{L_i(\cdot),L_{i+1}(\cdot),\ldots,L_{i+j}(\cdot)\}$ for some $j\geq 1$. Let $\{\lambda_i,\lambda_{i+1},\ldots,\lambda_{i+j}\}$ be some corresponding auxiliary parameters. Let the minimizer of the composite loss be 
\begin{align}
	y^*=\argmin \sum_{n=i}^{i+j}L_n(y).
\end{align}
Let there be a sequential update rule to determine $y^*$ and the new auxiliary variable $\lambda^*$ as
\begin{align}
	y^*=&F(y_i,y_{i+1},\ldots,y_{i+j};\lambda_i,\lambda_{i+1},\ldots,\lambda_{i+j}),\label{updateF}\\	\lambda^*=&G(y_i,y_{i+1},\ldots,y_{i+j};\lambda_i,\lambda_{i+1},\ldots,\lambda_{i+j})\label{updateG}.
\end{align}
Then, we can change the algorithm as follows:
\begin{enumerate}
	\item At the beginning, let us have $N$ samples with the corresponding scores $\{x_n\}_{n=1}^N$ and losses $\{l_n(\cdot)\}_{n=1}^N$.
	\item Create a group for each sample, i.e., $n\in\{1,\ldots,N\}$. Set the initial minimizers $y_n=\argmin_y l_n(y)$  and the auxiliary variables $\lambda_n$ for all $n\in\{1,\ldots,N\}$ groups.
	\item Let $\mathcal{I}$ be the set of all indices $i$ such that the consecutive group pair $(i,i+1)$ have $y_i\geq y_{i+1}$\label{step:checkE}
	\item If $\mathcal{I}$ is not empty, join the groups $i$ and $i+1$ for all $i\in\mathcal{I}$, and re-index the new groups accordingly. If there is consecutive grouping, we join all of them together. For example, if we join the pairs $\{n,n+1\}$ and $\{n+1,n+2\}$, we join $\{n,n+1,n+2\}$. 
	The new group losses and auxiliary variables are calculated in accordance with \eqref{updateF} and \eqref{updateG} respectively.
	Return to Step \ref{step:checkE}.
	\item ELSE (if $\mathcal{I}$ is empty), stop.
\end{enumerate}

\begin{remark}
	For the weighted square errors in \autoref{sec:square}; if we set the auxiliary variable as the total weight sum of the groups, we have
	\begin{align*}
		F(y_i,y_{i+1},\ldots,y_{i+j};\lambda_i,\lambda_{i+1},\ldots,\lambda_{i+j})=&\frac{\sum_{i'=i}^{i+j}\lambda_{i'}y_{i'}}{\sum_{i'=i}^{i+j}\lambda_{i'}},\\
		G(y_i,y_{i+1},\ldots,y_{i+j};\lambda_i,\lambda_{i+1},\ldots,\lambda_{i+j})=&{\sum_{i'=i}^{i+j}\lambda_{i'}},
	\end{align*}
	which takes $O(1)$ time to update the weighted means and the weight sums per step decrease (joining).
	Thus, this approach can find an optimal transform in $O(T)$ space and $O(T)$ time.
\end{remark}

\begin{example}\label{example2}
	For the same example in \autoref{example}, the efficient algorithm works as follows:
	\begin{enumerate}
		\item 
		$y:{44},{52},{18},{14},{93},{37},{96},{08},{01},{95},{21},{77},{46},{36},{69}$\\
		$\lambda:{01},{01},{01},{01},{01},{01},{01},{01},{01},{01},{01},{01},{01},{01},{01}$ 
		
		\item $y:{44},{28},{65},{35},{58},{53},{69}$\\
		$\lambda:{01},{03},{02},{03},{02},{03},{01}$ 
		
		\item $y:{32},{47},{55},{69}$\\
		$\lambda:{04},{05},{05},{01}$
		
		\item Since there are no more groups to combine, the algorithm stops and we reach the optimal staircase mapping.
	\end{enumerate}
\end{example}

\section{Online Algorithm to Find the Optimal Staircase Mapping for Strictly Convex Losses}\label{sec:online}

Suppose the samples come in an ordered fashion and the goal is to find the optimal staircase with each new sample.
Let the optimal transform for the first $N$ samples be $C_N(\cdot)$, then we have
\begin{align}
	\sum_{n=1}^Nl_n(C_N(x_n))=\min_{C(\cdot)\in\Omega}\sum_{n=1}^Nl_n(C(x_n)),
\end{align}
for all $N$.

Our aim is to sequentially update $C_N(\cdot)$ to find the next staircase $C_{N+1}(\cdot)$. We observe that the order in which we join the groups does not matter in \autoref{sec:offline}. Henceforth, the online algorithm works as follows:

\begin{enumerate}
	\item At the beginning, we have $N=1$, the score $x_1$, the minimizer $y_1$ and the auxiliary variable $\lambda_1$,
	\item Receive $x_{N+1}$, $y_{N+1}$, $\lambda_{N+1}$; and treat it as a new group with the appropriate index. \label{step:checkO}
	\item WHILE the largest indexed group should be joined with its immediate preceding group, join and re-index them (where the new index is the preceding group's index). The new group losses and auxiliary variables are calculated in accordance with \eqref{updateF} and \eqref{updateG} respectively.
	\item Set $N\leftarrow N+1$ and return to Step \ref{step:checkO}.
\end{enumerate}

\begin{example}
For the same example as in \autoref{example} and \autoref{example2}, the online algorithm works as the following where each step corresponds to the arrival of a new sample and the corresponding lines are the updates to the optimal staircase so far.
	\begin{enumerate}
		\item$44$
		\item$44,52$
		\item$44,52,18$\\
		$44,35,35$\\
		$38,38,38$
		\item$38,38,38,14$\\
		$32,32,32,32$
		\item$32,32,32,32,93$
		\item$32,32,32,32,93,37$\\
		$32,32,32,32,65,65$
		\item$32,32,32,32,65,65,96$
		\item$32,32,32,32,65,65,96,08$\\
		$32,32,32,32,65,65,52,52$\\
		$32,32,32,32,58.5,58.5,58.5,58.5$
		\item$32,32,32,32,58.5,58.5,58.5,58.5,01$\\
		$32,32,32,32,47,47,47,47,47$
		\item$32,32,32,32,47,47,47,47,47,95$
		\item$32,32,32,32,47,47,47,47,47,95,21$\\
		$32,32,32,32,47,47,47,47,47,58,58$
		\item$32,32,32,32,47,47,47,47,47,58,58,77$
		\item$32,32,32,32,47,47,47,47,47,58,58,77,46$\\
		$32,32,32,32,47,47,47,47,47,58,58,61.5,61.5$
		\item$32,32,32,32,47,47,47,47,47,58,58,61.5,61.5,36$\\
		$32,32,32,32,47,47,47,47,47,58,58,53,53,53$\\
		$32,32,32,32,47,47,47,47,47,55,55,55,55,55$
		\item$32,32,32,32,47,47,47,47,47,55,55,55,55,55,69$
	\end{enumerate}
	where the last step does not require any joining and is the same offline result.
\end{example}

At any time $T$, let the optimal transform have $S$ steps in the staircase. This approach has exactly $T-S$ combinations in total, which is in the worst-case $O(T)$. Thus, this approach can find an optimal transform in $O(T)$ space and $O(T\tau_{u})$ time, where $\tau_{u}$ is the total update time of \eqref{updateF} and \eqref{updateG}. 
\begin{remark}\label{rem:reg}
	For the weighted square error, $\tau_u=O(1)$, since when $i$ and $i+1$ groups are joined, we have
	\begin{align}
		y^*=\frac{\lambda_iy_i+\lambda_{i+1}y_{i+1}}{\lambda_i+\lambda_{i+1}}, &&\lambda^*=\lambda_i+\lambda_{i+1}.
	\end{align}
\end{remark}

\begin{remark}\label{rem:cla}
	For the weighted log-loss for binary classification problems, let $b_n\in\{0,1\}$ be the target binary class and $p_n$ be the probability of class $1$. The loss is
	\begin{align*}
		\min_{C(\cdot)\in\Omega_p}\sum_{n=1}^N-\alpha_n[b_n\log(C(p_n))+(1-b_n)\log(1-C(p_n))],
	\end{align*}
	where $\Omega_p$ is the class of transforms that map to $[0,1]$. The minimizers of this problem are the same as the following weighted square error problem
	\begin{align*}
		\min_{C(\cdot)\in\Omega_p}\sum_{n=1}^N\alpha_n(b_n-C(p_n))^2.
	\end{align*}
	Thus, the update rules are the same with \autoref{rem:reg}, where the initial conditions are $y_n=b_n$ and $\lambda_n=\alpha_n$; which results in $\tau_u=O(1)$.
\end{remark}

\begin{remark}
	Note that, if the samples come in an unordered fashion, we may not be able to produce an efficient algorithm for general strictly convex functions. To see this, consider the simple square loss. Suppose for some ordered $x_n$, we have $y_n=n$, where $\max_n y_n=N$. Since the target is also ordered, the optimal transform gives $C(x_n)=n$. Let the next sample have the minimum score $x$ so far, i.e., $x=\min_n x_n$ with a target $y=K$ for some large $K$. Then, the optimal transform will become a constant function $C_0(x_n)=\frac{K}{N+1}+\frac{N}{2}$. Suppose, we receive the score and target pairs $\{\min_n x_n,-K\}$, $\{\max_n x_n+1, y_{\argmax_n x_n}+1\}$ and $\{\min_n x_n,K\}$ in a periodic fashion. In this case, we will have an alternating optimal staircase function between constant and linear, which will have $O(T^2)$ complexity to update.
\end{remark}

\begin{remark}
	Note that although the cumulative time complexity of the online approach is $O(T\tau_u)$, the per sample time complexity can be up to $O(T\tau_u)$ by itself. However, we observe that although the per sample complexity is not efficiently bounded, the total time complexity is bounded because of the limited decrease in the number of steps of the staircase function.
\end{remark}

\begin{remark}
	In the online implementation of the algorithm, for unordered incoming samples, the approach in \cite{gokcesu2021optimally} is promising. However, unlike \cite{gokcesu2021optimally}, the complexity is not straightforward to bound because of the inherent difference in the optimal transform (staircase as opposed to the thresholding), which needs further analysis.  
\end{remark}

\section{Anytime Algorithm to Find the Optimal Staircase Mapping for Strictly Convex Losses}\label{sec:anytime}
We point out that while the minimization of relatively simple functions like the square loss in \autoref{sec:square} or the linear loss in \cite{gokcesu2021optimally} are straightforward; for general strictly convex losses, it can be costly. To this end, in this section, we propose an efficient anytime algorithm. 
Let the samples $x_n$ be ordered with individual losses $l_n(\cdot)$. 
The algorithm works as follows:
\begin{enumerate}
	\item At the beginning create $n$ groups with samples $x_n$. Set their group loss as $L_n(\cdot)=l_n(\cdot)$. Set the bounds on the minimizer $y_n$ as $A_n=A$ and $B_n=B$ for some $A,B$.
	\item For every group $i$ with $A_i\neq B_i$, evaluate the negative derivative at the middle of bounds $D_i=-L'_i(C_i)$, where $C_i=\frac{A_i+B_i}{2}$.\label{step:sample}
	\item WHILE there is a consecutive group pair $(i,i+1)$ such that $A_i=A_{i+1}$, $B_i=B_{i+1}$, and $D_i\geq 0\geq D_{i+1}$; join the groups $(i,i+1)$. The new group has $A_*=A_i$, $B_*=B_i$,
	$C_*=C_i$,
	$D_*=D_i+D_{i+1}$, $L_*(\cdot)=L_i(\cdot)+L_{i+1}(\cdot)$.\label{step:join}
	\item Re-index all groups. For every group $i$\label{step:D}
	\subitem IF $D_i\geq0$; $B_i=C_i$,
	\subitem IF $D_i\leq 0$; $A_i=C_i$.
	\item Return to Step \ref{step:sample}.
\end{enumerate}

At the end of any iteration of the algorithm, i.e., before returning to Step \ref{step:sample}, we have the following results.

\begin{proposition}\label{thm:A>B}
	For any group $i$, we have 
	\begin{align*}
		A_i>&B_i, \text{ if }D_i\neq 0;\\
		A_i=&B_i, \text{ if }D_i= 0.
	\end{align*}
	\begin{proof}
		Proof comes from Step \ref{step:sample} and Step \ref{step:D}.
	\end{proof}
\end{proposition}

\begin{proposition}\label{thm:D0}
	If a group $i$ has $D_i=0$, there is no other group $i'$ with the same bounds $A_i=B_i=A_{i'}=B_{i'}$.
	\begin{proof}
		The proof comes from Step \ref{step:join}.
	\end{proof}
\end{proposition}

\begin{proposition}\label{thm:DD}
	For any group pair $(i,i')$ with $A_i=A_{i'}$ and $B_i=B_{i'}$, we have
	\begin{align*}
		D_i>0 \text{ if }D_{i'}>0,\\
		D_i<0 \text{ if }D_{i'}<0.
	\end{align*} 
	\begin{proof}
		Let us have groups with $A_i=A'$ and $B_i=B'$ at some iteration $k$. Then, we will sample the derivatives at the same point $C_i=C'$. The resulting bounds are updated by either $A_i=C'$ if $D_i<0$ or $B_i=C'$ if $D_i>0$ and both if $D_i=0$ from Step \ref{step:D}. Thus, for any two groups to have the same upper and lower bounds, their $D_i$ needs to be non-zero and have the same sign.
	\end{proof}
\end{proposition}

\begin{lemma}\label{thm:ABA}
	For any group pair $(i,i+1)$, we either $A_{i+1}=A_i$ and $B_{i+1}=B_i$ or $A_{i+1}\geq B_{i+1}\geq A_i$.
	\begin{proof}
		The result comes from \autoref{thm:A>B}, \autoref{thm:D0} and \autoref{thm:DD}; where with each iteration the bound pair $(A_i,B_i)$ is progressively split in a monotone fashion.
	\end{proof}
\end{lemma}

\begin{lemma}\label{thm:0d}
	For any group $i$, where $A_i\neq B_i$; we have
	\begin{align*}
		A_i-B_i=\delta.
	\end{align*}
	where $\delta=2^{-k}(A-B)$ at the end of the $k^{th}$ iteration.
	\begin{proof}
		The proof comes from the fact that the split or halving the distance $A_i-B_i$ is definite and happens regardless. Only when $D_i=0$, this distance reduces to zero.
	\end{proof}
\end{lemma}

\begin{theorem}
	At the end of the $k^{th}$ iteration, we create $I$ groups with upper bounds $A_i$ and lower bounds $B_i$ such that they are ordered ($A_i\leq A_{i+1}$ and $B_i\leq B_{i+1}$) and $A_i-B_i\in\{0,\delta\}$, where $A=B+2^k\delta$.
	Moreover, $A_i\in\{B+\delta, B+2\delta,\ldots,B+2^k\delta\}$. When $A_i=B_i$, we have $D_i=0$. When $A_i\neq B_i$ and $A_i=B+N\delta$ for some $N$, we have
	\begin{align*}
		D_i>&0 \text{ if $N$ is even},\\
		D_i<&0 \text{ if $N$ is odd}.
	\end{align*}
	\begin{proof}
		The proof comes from \autoref{thm:ABA}, \autoref{thm:0d} and the progressive halving of the $A_i-B_i$ distance.
	\end{proof}
\end{theorem}

The space complexity is $O(T)$ since we keep the variables $A_i$, $B_i$, $D_i$ and the losses $L_i(\cdot)$. 

For the time complexity, we make at most $T$ number of evaluations $D_i$ at $C_i$ for each iteration, which takes $O(T)$ time per iteration. It takes $O(1)$ time for each joining of the groups. Thus, at the end of $k^{th}$ iteration, we take $O(kT)$ time. 

For a sensitivity of at least $\delta$ (i.e., height of the bounds $A_i-B_i$), we need $k\geq \log_2(\frac{A-B}{\delta})$. Thus, the time complexity is $O(T\log(1/\delta))$.
While this $\delta$ sensitivity can be preset as a stopping criterion, there exists an inherent sensitivity for only deciding the staircase regions. At any iteration, if there exist a group with unique $(A_i,B_i)$ pair, no further sampling will change this group and only increase the sensitivity of its minimizer. If the optimal staircase function has a minimum step of $\delta_0$, the time complexity is $O(T\log(1/\delta_0))$

If there is no explicit $A$, $B$ bounds available, we can use doubling trick schemes; where we sample the following points
\begin{itemize}
	\item When $A_i=\infty$ and $B_i=-\infty$, $C_i=0$.
	\item When $A_i=\infty$ and $B_i=0$, $C_i=1$.
	\item When $A_i=\infty$ and $B_i\geq 1$, $C_i=2B_i$.
	\item When $A_i=0$ and $B_i=-\infty$, $C_i=-1$.
	\item When $A_i\leq -1$ and $B_i=-\infty$, $C_i=2A_i$.
\end{itemize} 

Then, our algorithm works the same as before once the bounds $A_i$ and $B_i$ are finite, i.e., when $A_i<\infty$ and $B_i>-\infty$, $C_i=\frac{A_i+B_i}{2}$.

With this sampling we will incur an additional time complexity $O(T\log(\abs{A}+\abs{B}))$ of unknown $A$ and $B$.

\section{Conclusion}\label{sec:conc}
In this paper, we have investigated the optimal monotone transforms on score outputs that minimize certain losses. We started with the traditional square error setting and studied its weighted variant, where we showed that the optimal transform is in the form of a unique staircase function. We have further showed that this behavior is preserved for general strictly convex loss functions as well and showed that the optimizer transform is unique (hence, a global minimizer). We have analyzed the complexities of our approach for both offline and online scenarios, which creates the transform by successive partial minimizations. We have also extended our results to cases where the functions are not trivial to minimize. In such scenarios, we have proposed an anytime algorithm with increased precision per iteration.

\bibliographystyle{ieeetran}
\bibliography{double_bib}
\end{document}